\documentclass[10pt,twocolumn,letterpaper]{article}

\usepackage{cvpr}              
\definecolor{cvprblue}{rgb}{0.21,0.49,0.74}
\usepackage[pagebackref,breaklinks,colorlinks,allcolors=cvprblue]{hyperref}
\usepackage{algorithm}
\usepackage{algorithmicx}
\usepackage{algpseudocode}

\floatname{algorithm}{Algorithm}
\usepackage{amsmath}
\allowdisplaybreaks[4]
\usepackage{amssymb}
\usepackage{amsthm}
\usepackage{bm}
\usepackage{colortbl,xcolor} 
\usepackage{array,booktabs,multirow}
\usepackage{arydshln}  
\usepackage[table]{xcolor} 
\definecolor{darkgray}{gray}{0.85} 
\usepackage{booktabs}
\usepackage{multirow}
\usepackage{array}
\usepackage[table]{xcolor}
\usepackage{caption} 

\title{TreeQ: Pushing the Quantization Boundary of Diffusion Transformer via Tree-Structured Mixed-Precision Search}

\author{
  {Kaicheng Yang}$^{1}$,\enspace
  {Kaisen Yang}$^{2}$,\enspace
  {Baiting Wu}$^{2}$,\enspace \\
  ~{Xun Zhang}$^{1}$,\enspace 
  {Qianrui Yang}$^{2}$,\enspace 
  {Haotong Qin}$^{3}$,\enspace
  {He Zhang}$^{4}$,\enspace
  {Yulun Zhang}$^{1}$\thanks{Corresponding author: Yulun Zhang, yulun100@gmail.com}\\
  \textsuperscript{1}Shanghai Jiao Tong University,\enspace
  \textsuperscript{2}Tsinghua University,\enspace
  \textsuperscript{3}ETH Z\"{u}rich,\enspace
  \textsuperscript{4}Adobe Research\\
  \vspace{-10mm}
}

\usepackage{amssymb,amsthm}
\usepackage{amsmath}

\begin{document}
\maketitle

\begin{abstract}
Diffusion Transformers (DiTs) have emerged as a highly scalable and effective backbone for image generation, outperforming U-Net architectures in both scalability and performance. However, their real-world deployment remains challenging due to high computational and memory demands. Mixed-Precision Quantization (MPQ), designed to push the limits of quantization, has demonstrated remarkable success in advancing U-Net quantization to sub-4-bit settings while significantly reducing computational and memory overhead. Nevertheless, its application to DiT architectures remains limited and underexplored. In this work, we propose \textbf{TreeQ}, a unified framework addressing key challenges in DiT quantization. First, to tackle inefficient search and proxy misalignment, we introduce \textbf{Tree-Structured Search (TSS)}. This DiT-specific approach leverages the architecture's linear properties to traverse the solution space in $\mathcal{O}(n)$ time while improving objective accuracy through comparison-based pruning. Second, to unify optimization objectives, we propose \textbf{Environmental Noise Guidance (ENG)}, which aligns Post-Training Quantization (PTQ) and Quantization-Aware Training (QAT) configurations using a single hyperparameter. Third, to mitigate information bottlenecks in ultra-low-bit regimes, we design the \textbf{General Monarch Branch (GMB)}. This structured sparse branch prevents irreversible information loss, enabling finer detail generation. Through extensive experiments, our \textbf{TreeQ} framework demonstrates state-of-the-art performance on DiT-XL/2 under W3A3 and W4A4 PTQ/PEFT settings. Notably, our work is the first to achieve near-lossless 4-bit PTQ performance on DiT models. The code and models will be available at \url{https://github.com/racoonykc/TreeQ}.
\end{abstract}

\setlength{\abovedisplayskip}{2pt}
\setlength{\belowdisplayskip}{2pt}

\vspace{-4mm}
\section{Introduction}
\vspace{-2mm}
Model quantization has emerged as a key approach to mitigating the computational bottlenecks of diffusion models. Within existing quantization paradigms, particularly Post-Training Quantization (PTQ)~\cite{shang2023post, li2023qdiffusion, he2023ptqd, so2023temporaldynamicquantizationdiffusion, tang2024post, huang2023tfmq, wu2024ptq4dit} and Quantization-Aware Training (QAT)~\cite{li2023qdm, zheng2025binarydm, zheng2024bidm, wang2025quest, he2023efficientdm, feng2025mpq, li2024svdquant, feng2025mpqdm2}, allocating appropriate precision is critical. Uniform precision quantization often fails to accommodate the diverse activation distributions in diffusion models, which vary significantly across layers. This heterogeneity introduces redundancy: excessive precision for insensitive components wastes resources, while insufficient precision for sensitive components degrades performance.

\begin{figure}[t!]
\centering
\vspace{-5mm} 
\captionsetup[subfigure]{labelformat=empty}

\begin{subfigure}[b]{0.49\linewidth} 
    \centering

\includegraphics[width=\linewidth]{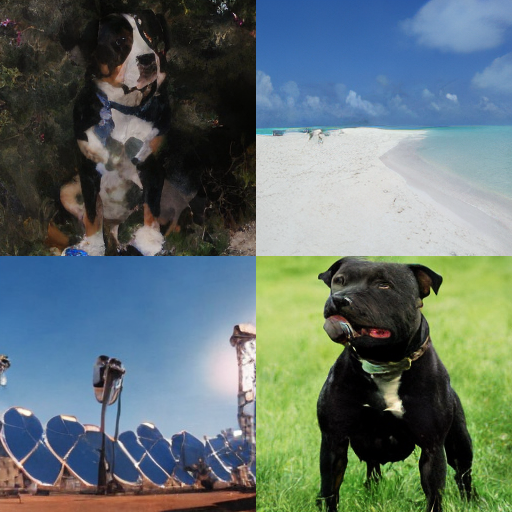}

    \vspace{-1mm} 
    \caption{W4A4}
    \label{fig:vis-tl}
\end{subfigure}
\hfill 
\begin{subfigure}[b]{0.49\linewidth} 
    \centering

\includegraphics[width=\linewidth]{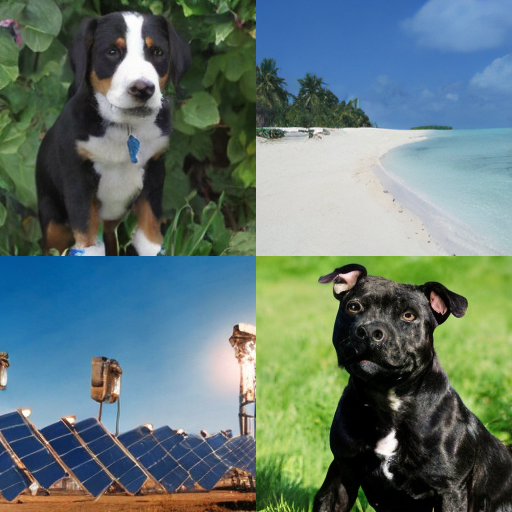}

    \vspace{-1mm} 
    \caption{W4A4}
    \label{fig:vis-tr}
\end{subfigure}

\vspace{-1mm}

\begin{subfigure}[b]{0.49\linewidth}
    \centering

\includegraphics[width=\linewidth]{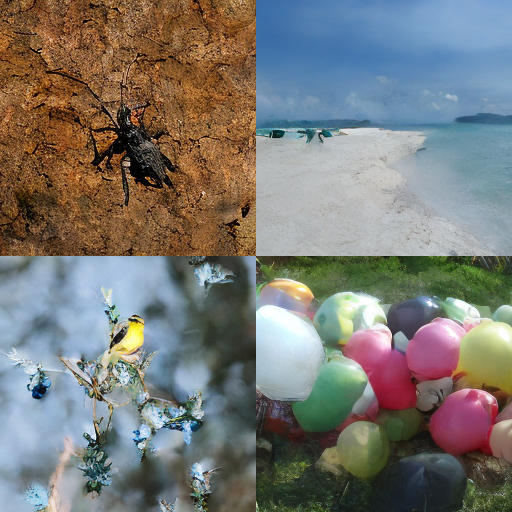}

    \vspace{-1mm} 
    \caption{W3A3}
    \label{fig:vis-bl}
\end{subfigure}
\hfill
\begin{subfigure}[b]{0.49\linewidth} 
    \centering

\includegraphics[width=\linewidth]{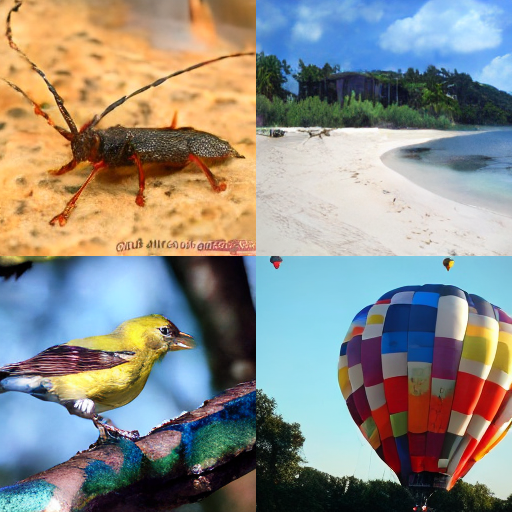}

    \vspace{-1mm} 
   \caption{W3A3}
    \label{fig:vis-br}
\end{subfigure}

\vspace{-4.5mm}
\caption{Visual comparison on DiT-XL/2 under low-bit PTQ. TreeQ achieves better generation compared to baseline~\cite{yang2025robuq}.}
\vspace{-8mm} 
\label{fig:visualization} 
\end{figure}

Mixed-precision quantization (MPQ) addresses these issues by allocating different bit-widths to different model components, enabling flexible target average bit-widths and improved performance. However, existing MPQ approaches face major challenges: (i) proxy objectives often misalign with true performance metrics, leading to suboptimal assignments~\cite{shao2025tr, zhao2024mixdq, feng2025mpq, feng2025mpqdm2}, (ii) search procedures are computationally expensive~\cite{lee2025amq, mills2024qua2sedimo}, and (iii) in extreme low activation bit regimes (e.g., 3-bit), MPQ suffers from propagation bottlenecks that cannot be recovered by simple low-rank auxiliary branches.

To overcome these challenges, we propose TreeQ, a unified framework addressing each of these issues. 
First, to tackle inefficient search (ii) and proxy misalignment (i), we propose Tree-Structured Search (TSS), a search framework designed for DiT architectures. It leverages the linear properties of the DiT structure to efficiently traverse the mixed-precision solution space with $\mathcal{O}(n)$ time complexity and improves search objective accuracy through a comparison-based pruning method.
Second, to further address the objective misalignment (i) between different quantization schemes, we introduce Environmental Noise Guidance (ENG). This method guides the optimization objective within the TSS framework to favor either PTQ or QAT via a simple hyperparameter.
Third, to break the propagation bottlenecks (iii) in low-bit regimes, we introduce the General Monarch Branch (GMB). This structured sparse matrix branch overcomes irreversible information loss caused by low-bit quantization and low-rank branches, enabling finer detail generation.
Extensive experiments and ablation studies demonstrate the effectiveness of each component, which collectively advance the state-of-the-art towards lower-bit and more practical DiT models.

Our main contributions are:
\begin{itemize}
    \item We introduce \textbf{TSS}, a search framework designed for DiT architectures that efficiently explores the mixed-precision solution space with $\mathcal{O}(n)$ time complexity for $n$-layer models. It enhances search objective accuracy through comparison-based bottom-up merging and pruning.
    
    \item We introduce \textbf{ENG}, which guides the optimization objective within the TSS framework to favor either PTQ or QAT via a simple hyperparameter.

    \item We introduce \textbf{GMB}, a structured sparse matrix branch that overcomes irreversible information loss caused by low-bit quantization and standard low-rank branches, enabling finer detail generation in quantized models.

    \item Our proposed framework, \textbf{TreeQ}, achieves state-of-the-art performance on DiT-XL/2 under W3A3 and W4A4 PTQ/PEFT settings, and is the first to achieve near-lossless 4-bit PTQ performance on DiT-XL/2.
\end{itemize}
\vspace{-1mm}
\section{Related Work}
\noindent{\textbf{Quantization of Diffusion Models.}}
Diffusion Models (DMs) have demonstrated strong generative capabilities across vision and multimodal tasks~\cite{chen2020wavegrad,hu2022st,rombach2022high,chen2023hierarchical,he2023reti,li2023mhrr,li2023mpgraf,liu2024intelligent,li2024snapfusion,he2024diffusion,ho2020denoising,zhao2024dcsolver,peebles2023scalable}. To improve scalability and generality, recent works replace the traditional U-Net~\cite{ronneberger2015u} with Transformer architectures~\cite{vaswani2017attention}~\cite{croitoru2023diffusion,rombach2022high,yang2023diffusion}. Diffusion Transformers (DiTs)~\cite{peebles2023scalable} achieve strong performance; however, their inference remains expensive in both memory and computation, limiting practical deployment.
\begin{figure}[t!]
    \centering
    \includegraphics[height=5.3cm]{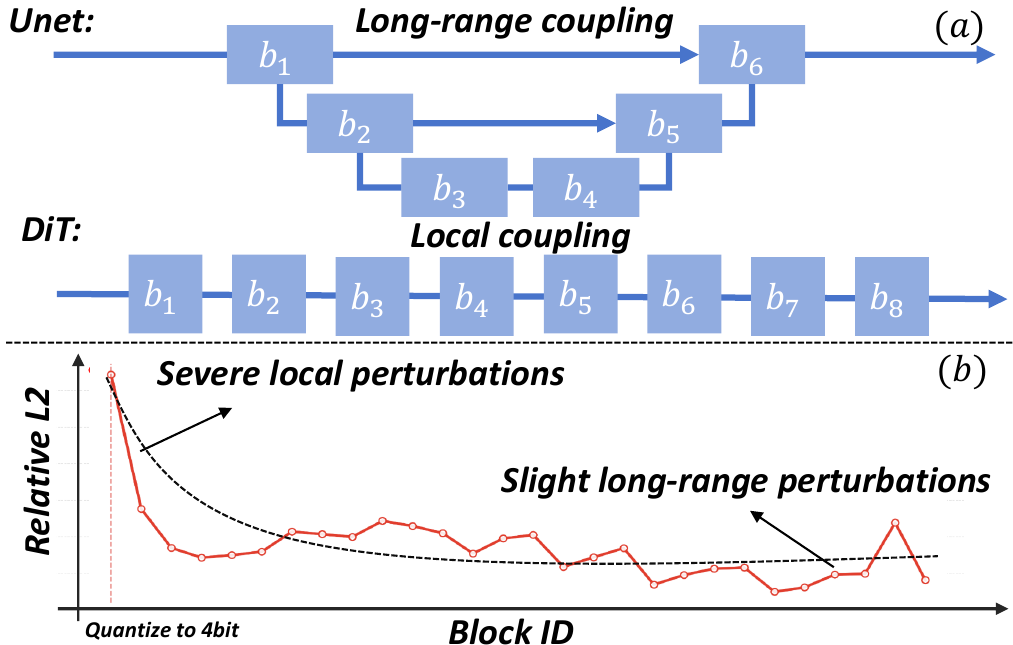}
    \vspace{-6mm}
    \caption{Analysis of DiT's local coupling properties. (a) U-Net's skip connections create long-range dependencies, whereas DiT exhibits a simple linear structure of block connections. (b) Relative L2 loss versus distance when quantizing the first block's attn.qkv layer to 4-bit. This inspires our tree-structured merging algorithm, which progressively aggregates local segments into global configurations while ensuring evaluation of each adjacency relationship.}
    \vspace{-5mm}
    \label{fig:analy}
\end{figure}
To reduce inference cost, post-training quantization (PTQ) methods~\cite{shang2023post,li2023qdiffusion} reconstruct activations using calibration, while subsequent work improves temporal robustness and efficiency~\cite{he2023ptqd,so2023temporaldynamicquantizationdiffusion,tang2024post}. Transformer-based DMs further adopt salience-guided calibration~\cite{huang2023tfmq,wu2024ptq4dit}. A core challenge remains handling outliers: methods such as SVDQuant~\cite{li2024svdquant} and EfficientDM~\cite{he2023efficientdm} mitigate them using LoRA, whereas TRDQ~\cite{shao2025tr}, RobuQ~\cite{yang2025robuq}, and HadaNorm~\cite{federici2025hadanorm} leverage orthogonal transforms. On the quantization-aware training (QAT) side, Q-dm~\cite{li2023qdm}, BinaryDM~\cite{zheng2025binarydm}, and BiDM~\cite{zheng2024bidm} achieve ultra-low bit regimes, while QuEST~\cite{wang2025quest} and EfficientDM~\cite{he2023efficientdm} reduce fine-tuning costs.

\noindent{\textbf{Mixed-Precision Quantization.}}
Mixed-precision quantization (MPQ) assigns heterogeneous bit-widths to model components to reflect non-uniform sensitivity under resource constraints. In large language models (LLMs), AMQ~\cite{lee2025amq} uses AutoML for layer-wise bit selection, while LLM-MQ adopts gradient-based sensitivity with constrained optimization. For Vision Transformers, FIMA-Q~\cite{wu2025fima} and APHQ-ViT~\cite{wu2025aphq} employ Fisher-based or Hessian-based sensitivity metrics to guide low-bit reconstruction. For diffusion models, methods such as Mix-DQ~\cite{zhao2024mixdq}, MPQ-DM~\cite{feng2025mpq}, MPQ-DMv2~\cite{feng2025mpqdm2}, Mix-DiT~\cite{kim2025mixdit} and Qua2SeDiMo~\cite{mills2024qua2sedimo} investigate mixed-precision planners based on integer programming and GNN-based search. These studies suggest that adaptive bit-width assignment is effective for maintaining performance at low bit setting.

\noindent{\textbf{Structured Sparse Matrices.}}
Structured sparse matrices aim to achieve both hardware efficiency and parameter reduction on modern accelerators~\cite{zhang2024fine,zhuo2005sparse,zhu2013accelerating,feldmann2023spatula,dao2019learning}. Early work in this direction includes Butterfly matrices~\cite{li2015butterfly}, which are highly expressive but can be hardware-inefficient in practice. To address this limitation, hardware-aware variants such as Pixelated Butterfly~\cite{dao2021pixelated} were subsequently developed . A prominent example in this line of work is the Monarch matrix~\cite{dao2022monarch}, parameterized as a product of two block-diagonal matrices. This structure maps efficiently to hardware-native BMM routines~\cite{fu2023simple,fan2022adaptable,liu2023hardsea} and, importantly, enables optimal initialization from pre-trained dense weights via an analytical projection.
\vspace{-2mm}
\section{Methodology}
\begin{figure*}[t!]  
\centering
\includegraphics[height=8.25cm]{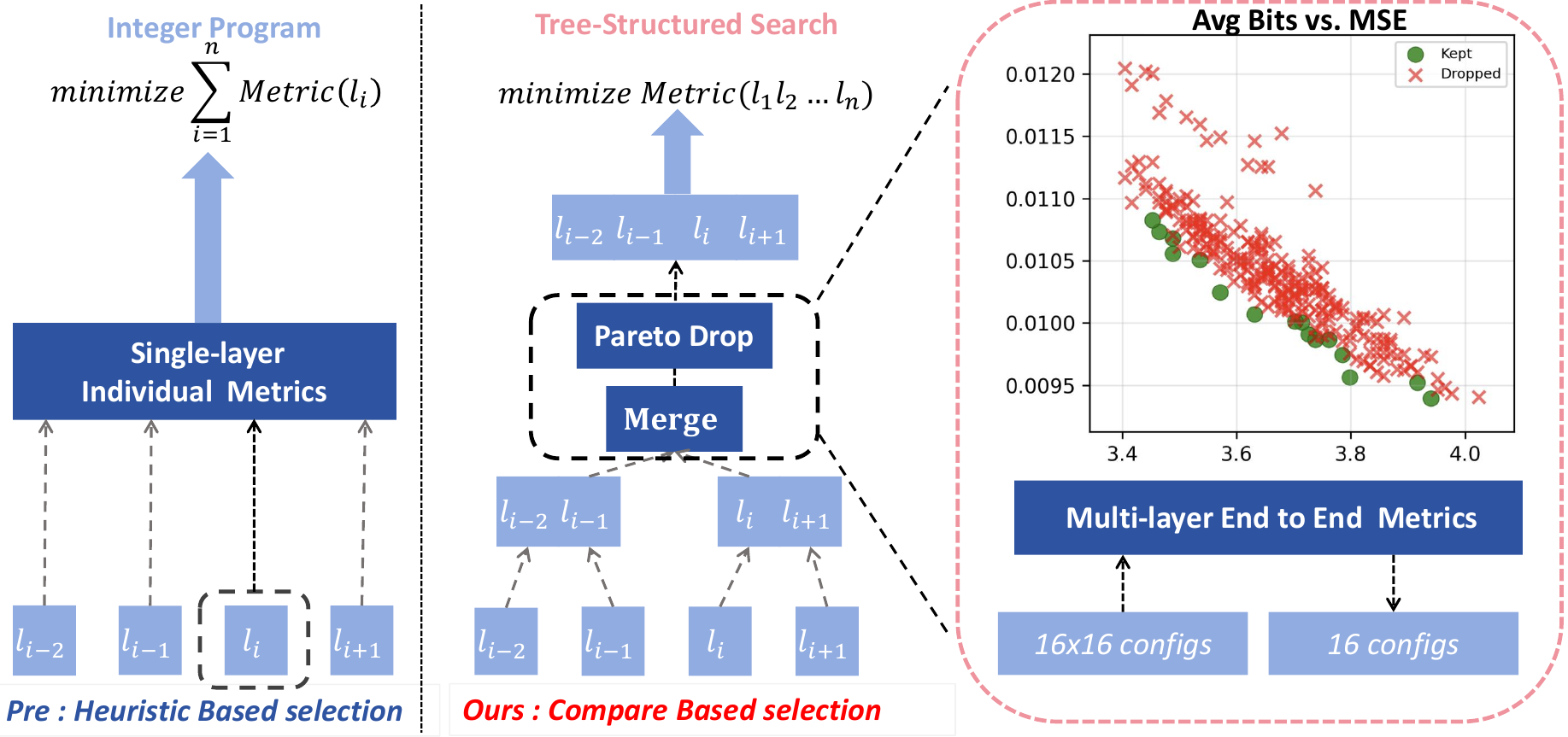}
\vspace{-6mm}
\caption{Visualization comparison between TSS and traditional methods Integer Programming. \textbf{(left)} Integer programming methods typically define a layer-wise heuristic function and set the optimization goal as minimizing the sum of these functions. This implies a necessary linearity between the heuristic and final performance, leading to an optimization objective error. \textbf{(right)} TSS leverages DiT's local coupling, merging the search strategy from local to global. This naturally considers the interactions between multiple layers and allocates more resources to strongly coupled adjacent layers. By adopting a comparison-based Pareto queue pruning strategy, the chosen objective function (e.g., MSE) only needs to be strongly order-preserving with the final performance, correcting the optimization objective.}

\label{fig:overview}
\vspace{-4.5mm}
\end{figure*}
\vspace{-1mm}
\subsection{Preliminary}
\label{sec:preliminary}
\vspace{-2mm}
In this section, we first describe the baseline quantization procedure \cite{yang2025robuq} used in our framework, which has demonstrated state-of-the-art (SOTA) performance.

\noindent{\textbf{Activation Quantization.}}
Let $\mathbf{X}\in\mathbb{R}^{T\times C}$ be token activations. Applying $\mathbf{H}_C$ yields $Y_{t,c}=\sum_{j}(\mathbf{H}_C)_{j,c}X_{t,j}$.
By generalized CLT, $Y_{t,\cdot}$ approaches $\mathcal{N}(0,\sigma_t^2)$, where
$\sigma_t^2=\frac{1}{C}\sum_j\text{Var}(X_{t,j})$.
The transformed activations are then quantized in the Gaussian domain using a high-bit \textbf{symmetric} uniform quantizer $Q_{\mathrm{uni}}(\cdot)$. This quantizer utilizes a pre-computed optimal  $\delta$, determined for the standard normal distribution $\mathcal{N}(0,1)$, and is designed for efficient deployment on INT operators:
\begin{equation}
Q_G(\mathbf{x})
=\sigma_t\cdot
Q_{\mathrm{uni}}\!\left(\frac{\mathbf{H}_n^{\top}\mathbf{x}}{\sigma_t}\right).
\end{equation}
\noindent{\textbf{Weight Quantization.}}
Given a weight matrix $\mathbf{W}\in\mathbb{R}^{m\times n}$, we first apply a normalized Hadamard transform $\mathbf{H}_n$ to obtain the transformed weight $\mathbf{W}_H=\mathbf{W}\mathbf{H}_n$, and perform SVD on it while explicitly retaining a rank-$r$ ($r{=}16$) full-precision component:
\begin{equation}
\mathbf{W}_H \approx \mathbf{A}\mathbf{B} = \mathbf{U}_r\boldsymbol{\Sigma}_r\mathbf{V}_r^{\top}.
\end{equation}
The residual $\mathbf{W}_\text{res} = \mathbf{W}_H - \mathbf{A}\mathbf{B}$ is then quantized channel-wise, leveraging the same symmetric uniform quantizer $Q_{\mathrm{uni}}(\cdot)$ from activation quantization:
\begin{equation}
Q_w(\mathbf{W}_\text{res}) = \sigma_c \cdot Q_{\mathrm{uni}}\!\left(\frac{\mathbf{W}_\text{res}}{\sigma_c}\right).
\end{equation}
where $\sigma_c$ is the channel-wise standard deviation of the residual $\mathbf{W}_\text{res}$.
The final reconstructed weight is:
\begin{equation}
\hat{\mathbf{W}} = (\underbrace{\mathbf{A}\mathbf{B}}_{\text{LRB}} + \underbrace{Q_w(\mathbf{W}_\text{res}))}_{\text{Quantized  \ Residual}}\mathbf{H}_n^{\top}.
\end{equation}
This yields an orthogonal-transform pipeline for both weights and activations, enabling stable and hardware-ready mixed-precision quantization.

\subsection{Tree-Structured Search (TSS)}
\label{sec:treeq-ptq}
\vspace{-1mm}
\subsubsection{Main Issues in Current Methods}
\label{sec:analysis}
We found that there are several remaining unsolved issues in current mixed-precision quantization for DiTs:

\begin{itemize}
\item \textbf{Under-utilization of DiT Topology.} Unlike U-Nets that contain global shortcut structures, DiTs are strictly linear. As shown in Fig.~\ref{fig:analy}, neighboring blocks in DiT exhibit much stronger mutual dependencies. Current methods fail to leverage this structural prior.
\item \textbf{Objective Misalignment.} Most existing layer-wise heuristic methods define sensitivity at the layer level, and then search configurations by minimizing a heuristic objective under a target bit budget. However, such heuristics ignore inter-layer interactions, resulting in objective misalignment and severe suboptimality in PTQ.
\item \textbf{Excessive Complexity.} ML-based configuration search requires auxiliary models and curated training data, thus complex, difficult to implement, and non-generalizable; enumeration or evolutionary approaches depend on high-quality metrics and are computationally prohibitive.
\end{itemize}

Motivated by these limitations, we aim to develop an algorithm that (i) fully exploits the locality induced by DiT’s linear topology (ii) operates via comparisons to eliminate reliance on expensive metrics and avoid objective misalignment (iii) achieves controllable time complexity that scales gracefully with model depth.

\subsubsection{Formal Description for TSS}
\noindent{\textbf{Definitions.}}
The DiT backbone is an ordered sequence of weights $\{w_i\}_{i=1}^N$. Any consecutive subsequence $m=\{w_i\}_{i=p}^q$ is a \textit{module}.

A \textit{configuration} $c_I$ for module $m$ with index set $I$ specifies the quantization bitwidth for each weight in the module, denoted as $c_I=\{b_i\}_{i\in I}$ where $b_i$ is the bitwidth assigned to weight $w_i$. The \textit{mean bitwidth} $\bar{c}(c_I)$ is defined as the FLOPs-weighted average across all weights in the module.

An \textit{environment} \(e\) specifies default bitwidths for layers outside \(m\). Given a configuration \(c_{m}\) under environment \(e\), \(p_e(c_{m})\) denotes a real-valued \textit{performance indicator}.

For a candidate set \(S_m=\{c_j\}_{j=1}^K\) on module \(m\), the \textit{Pareto queue} \(P_{(p,e)}(S_m)\) consists of non-dominated configurations with respect to \((p,\bar{c})\):
\begin{gather}
c \in P_{(p,e)}(S_m) 
\iff 
\neg \exists c' \in S_m \nonumber \\
\text{s.t. } 
\bigl[p_e(c') \le p_e(c)\bigr] \wedge \bigl[\bar{c'} \le \bar{c}\bigr].
\end{gather}
where at least one of the inequalities is strict.

If two queues correspond to disjoint modules \(m_1, m_2\), their \textit{Cartesian} is defined as
\begin{gather}
P_{(p,e)}(S_{m_1}) \otimes P_{(p,e)}(S_{m_2}) 
:= \nonumber \\
P_{(p,e)} \bigl(\{c_i \cup c_j \mid c_i \in S_{m_1},\, c_j \in S_{m_2}\}\bigr).
\end{gather}

Notably, this procedure relies only on relative comparisons of local structures, requiring the performance indicator to preserve the local ordering of configurations. This allows the use of inexpensive calibration MSE.

\noindent{\textbf{TSS procedure.}}
Figure~\ref{fig:overview} illustrates the hierarchical pipeline.
We first treat each layer as a node and enumerate all admissible bitwidths to form its local Pareto queue. Following DiT’s linear topology, we iteratively merge \emph{adjacent} nodes: for each neighboring pair, we take the Cartesian of their Pareto queues. As shown in Fig.~\ref{fig:overview}, the Cartesian product remains extremely sparse:
\begin{equation}
|P_{(p_1,e)}(S_{m_1}) \otimes P_{(p_2,e)}(S_{m_2})|
\ll
|S_{m_1}| \cdot |S_{m_2}|,
\end{equation}
ensuring that queue sizes remain small during bottom-up merging. If a merged queue exceeds the maximum size \(k\), we retain only the top \(k\) candidates whose mean bitwidths are closest to the target. This process continues until a single root queue remains, representing globally non-dominated candidates. The final configuration is chosen as the one whose mean bitwidth \(\bar{c}\) is closest to the target \(c_\text{target}\).

In Supplementary Material, we show that if a single indicator evaluation costs \(\alpha\), the Pareto queue size is \(k\), and the model has \(n\) layers, then TSS runs in \(\mathcal{O}(\alpha k^2 n)\), achieving controllable linear scaling with model depth. Notably, exactly \(n-1\) merging steps occur, corresponding precisely to the \(n-1\) local connections in the original DiT, naturally accounting for its strong local dependencies.

\begin{figure}[t!]
    \centering
    \includegraphics[height=6.0cm]{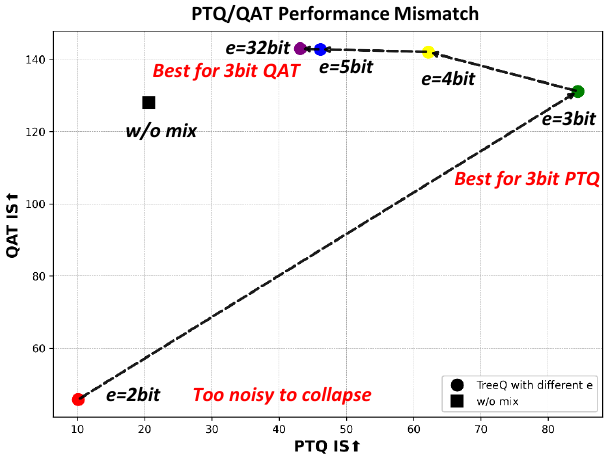}
    \vspace{-3mm}
    \caption{Visualization of ENG at 3-bit. We observe that the same mixed-precision configuration performs differently under PTQ and QAT, with some configurations favoring PTQ and others QAT. The environmental noise parameter $e$ is a highly sensitive hyperparameter that guides the search process. When applying excessive noise ($e$=2bit, below the target), the search collapses. However, setting $e$=3bit (matching the target) yields a configuration optimal for PTQ, whereas $e$=32bit (no noise) finds a configuration that adapts better to QAT.}

    \vspace{-10mm}
    \label{fig:analy_treeq2}
\end{figure}
\vspace{-2mm}

\begin{figure*}[t]  
\centering
\includegraphics[height=3.8cm]{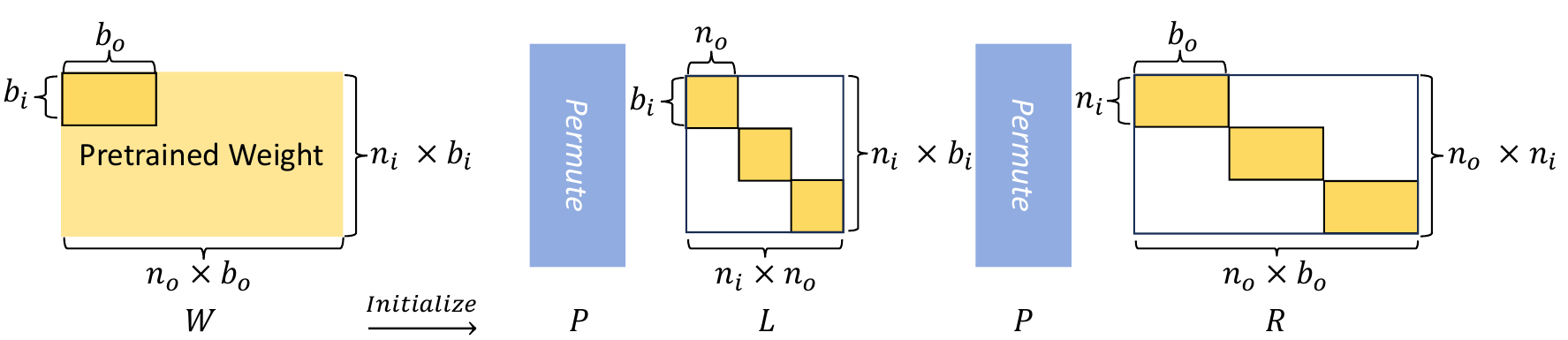}
\vspace{-7mm}
\caption{Visualization of GMB. GMB can initialize a structured sparse matrix with arbitrary sparsity from a pre-trained weight of any shape. The block-diagonal form of the sub-matrices facilitates efficient parallel processing on modern GPUs.}

\label{fig:skip connection}
\vspace{-4.5mm}
\end{figure*}

\subsubsection{Environment Noise Guidance}
\label{sec:adpatation}
\vspace{-1mm}
\noindent{\textbf{PTQ vs. QAT objectives.}}
PTQ and QAT have distinct objectives. For PTQ, the model cannot fine-tune away errors, so each layer must accommodate the macroscopic noise introduced by others. Traditional integer programming (IP) methods, however, often evaluate layers in a noise-free environment. While this isolates layer sensitivity, it ignores the inter-layer error propagation that causes catastrophic failure in PTQ. This noise-free evaluation is actually more suitable for QAT, where gradient-based fine-tuning aligns the quantized layers back toward full precision.

\noindent{\textbf{Unifying Objectives via Environment Noise Guidance.}}
To unify these objectives, we introduce \textbf{Environment Noise Guidance (ENG)} within the TSS framework. ENG uses a simple hyperparameter \(e\) to simulate the target context. As shown in Fig.~\ref{fig:analy_treeq2}, this parameter is highly sensitive. For PTQ, we inject realistic noise by setting \(e\) to the target bit-width, forcing the search to find a robust strategy. For QAT, we minimize noise ($e$=32bit) to find strategies best suited for fine-tuning. This flexibility in adapting to distinct quantization objectives (PTQ and QAT) via a single parameter, $e$, underscores the versatility design of the TSS framework.

\vspace{-2mm}
\subsection{General Monarch Branch}
\label{sec:GMB}
\vspace{-1mm}
\subsubsection{Issue of Local High-Frequency Information}
\vspace{-1mm}
In many low-bit quantized diffusion models, we observe a consistent phenomenon: global metrics such as FID~\cite{heusel2017gans} often remain reasonable, while spatial FID (sFID)~\cite{nash2021generating}, which evaluates local structure fidelity, deteriorates significantly. This discrepancy becomes more pronounced in ultra-low-bit quantization (e.g., 2-bit or 3-bit), indicating that while global, low-frequency information is largely preserved, fine-grained high-frequency details are substantially degraded. Such local information loss can adversely impact visual fidelity, especially for structured features critical for downstream perception.
A straightforward remedy is to increase the rank of low-rank branches. However, this approach is counterproductive for quantization: since parameter count grows linearly with rank, higher rank requires substantially much more parameters, directly opposing the goal of model compression. Alternatively, one may introduce sparse high-rank skip connections, but existing variants are typically \emph{unstructured} (e.g., top-$k$ element selection)~\cite{liu2025bimacosr} and misaligned with hardware-native block BMM kernels.

\subsubsection{General Monarch Matrices}
This motivates another direction: introducing a parallel \emph{structured sparse high-rank branch}, which is (i) \textbf{high-rank} enough to inject the missing local residuals, and (ii) \textbf{block-structured} to inherit the same kernel efficiency as standard block BMM. We take inspiration from Monarch matrices~\cite{dao2022monarch}, and generalize the original \emph{square-only} design to a block-partitioned form applicable to arbitrary weight shapes, yielding the \textbf{General Monarch Branch (GMB)}.

\subsubsection{General Monarch Decomposition}
Let $A \in \mathbb{R}^{N_o \times N_i}$ be a weight matrix. We first partition the input and output dimensions into blocks:
\begin{equation}
N_i = n_i \cdot b_i, \quad N_o = n_o \cdot b_o,
\end{equation}
defining $n_o \times n_i$ sub-matrices
\begin{equation}
M_{jk} \in \mathbb{R}^{b_o \times b_i}, \quad j=1,\dots,n_o, \ k=1,\dots,n_i.
\end{equation}
The full matrix can be reconstructed exactly as
\begin{equation}
A = \mathrm{concat}_{j,k} M_{jk},
\end{equation}
where $\mathrm{concat}_{j,k}$ denotes concatenation along rows and columns.

Each block $M_{jk}$ is then approximated by a rank-1 decomposition
\begin{equation}
M_{jk} \approx u_{jk} v_{jk}^\top, \quad
u_{jk} \in \mathbb{R}^{b_o}, \ v_{jk} \in \mathbb{R}^{b_i}.
\end{equation}
Aggregating all rank-1 factors constructs the core Monarch tensors by grouping the components dimension-wise:
\begin{equation}
L'[j,k,:] = u_{jk} \in \mathbb{R}^{b_o}, \quad 
R'[j,k,:] = v_{jk} \in \mathbb{R}^{b_i},
\end{equation}
for $j=1,\dots,n_o$ and $k=1,\dots,n_i$.
Using the Monarch permutation matrix $P$~\cite{dao2022monarch}, we transform this representation into a more efficient and compact factored form:
\begin{equation}
\hat{M} = P L P R,
\end{equation}
where the factor matrices $L \in \mathbb{R}^{ n_o b_o\times n_i n_o }$ and $R \in \mathbb{R}^{n_i n_o\times n_i b_i}$ are obtained by reshaping and permuting $L'$ and $R'$ to expose their block-diagonal structure, as illustrated in Fig.~\ref{fig:skip connection}, which provides a clear visualization.

To match the parameter budget of a low-rank branch $\mathbf{AB}$ with rank $r$, the parameter counts must satisfy:
\begin{equation}
n_in_o(b_i + b_o) = r(n_ob_o + n_ib_i).
\end{equation}
A simple and valid solution to this constraint is to set $n_i = n_o = r$. Consequently, in our experiments, we enforce this configuration where both the row and column partition counts are set to $r$ to achieve effective sparsity control while maintaining computational tractability.

\vspace{-2mm}
\subsubsection{Quantization with GMB}
\label{sec:gmb-quant}
Let $\mathbf{W}_H = \mathbf{W}\mathbf{H}_n$ denote the Hadamard-transformed weight matrix.  
In the transformed domain, $\mathbf{W}_H$ is decomposed into LRB, GMB and quantized residual:
\begin{equation}
\mathbf{W}_\text{res} = \mathbf{W}_H - \mathbf{A}\mathbf{B} - \mathbf{P L P R},
\end{equation}
\begin{equation}
\mathbf{\hat{W}}_H = \underbrace{\mathbf{A}\mathbf{B}}_{\text{LRB}}
+ \underbrace{\mathbf{P L P R}}_{\text{GMB}}
+ \underbrace{Q_w(\mathbf{W}_\text{res})}_{\text{Quantized Residual}}.
\end{equation}

The final reconstructed weight is obtained via the inverse transform:
\begin{equation}
\mathbf{\hat{W}} = \mathbf{\hat{W}}_H \mathbf{H}_n^\top.
\end{equation}

The corresponding forward computation is
\begin{equation}
y = Q_w(\mathbf{W}_\text{res}) Q_G(x)
   + (\mathbf{A}\mathbf{B} + \mathbf{L'R'})\mathbf{H}_n^\top x.
\end{equation}

\newcolumntype{C}[1]{>{\centering\arraybackslash}p{#1}}
\begin{table*}[t]
  \caption{Performance on ImageNet-1K $256{\times}256$ under different settings. Best results per bit-width are highlighted.}
  \vspace{-3mm}
  \label{tab:main_experiments_final}
  \footnotesize
  \centering
  \setlength{\tabcolsep}{3.0mm}
  \setlength{\arrayrulewidth}{0.1mm}
  \renewcommand{\arraystretch}{1.1}
  \resizebox{0.87\linewidth}{!}{
  \begin{tabular}{C{1.8cm} C{1.5cm} C{2.2cm} C{1.4cm} C{1.0cm} C{1.0cm} C{1.1cm} C{1.1cm}}
    \toprule[0.15em]
    Setting & Paradigm & Method & W/A & IS$\uparrow$ & FID$\downarrow$ & sFID$\downarrow$ & Precision$\uparrow$ \\
    \midrule

    \multirow{18}{=}{\centering ImageNet\\CFG = 1.5\\steps = 50}
      & \multirow{11}{=}{\centering PTQ} 
      & FP               & 32/32 & 241.18 & 6.28  & 20.78 & 0.783 \\
      \cdashline{3-8}
      & & SVD-Quant~\cite{li2024svdquant}       & 4/4   & 2.48   & 269.83 & 146.63 & 0.0070 \\
      & & PTQ4DiT~\cite{wu2024ptq4dit}         & 4/4   & 3.05   & 231.80 & 106.42 & 0.1003 \\
      & & Q-DiT~\cite{chen2024qdit}           & 4/4   & 2.01   & 248.11 & 404.44 & 0.0138 \\
      & & Quarot~\cite{ashkboos2024quarot}          & 4/4   & 53.12  & 53.31  & 56.74  & 0.4134 \\
      & & Baseline~\cite{yang2025robuq}           & 4/4   & 142.14 & 16.64 & 28.54 & 0.6446 \\
      & &  \textbf{TreeQ-PTQ} &  4/4 &  \textbf{219.66} &  \textbf{6.92} &  \textbf{20.86} &  \textbf{0.7664} \\
      \cdashline{3-8}
      & & SVD-Quant~\cite{li2024svdquant}       & 3/3   & 1.25   & 370.14 & 393.81 & 0.0000 \\
      & & Quarot~\cite{ashkboos2024quarot}          & 3/3   & 3.43   & 306.77 & 228.16 & 0.0171 \\
      & & Baseline~\cite{yang2025robuq}           & 3/3   & 20.59  & 99.99  & 67.54  & 0.2365 \\
      & &  \textbf{TreeQ-PTQ} &  3/3 &  \textbf{85.21} &  \textbf{28.08} &  \textbf{28.97} &  \textbf{0.5346} \\
      \cline{2-8}
      & \multirow{6}{=}{\centering PEFT}
      & QueST~\cite{wang2025quest}           & 4/4   & 4.87   & 215.06 & 72.15 & 0.0529 \\
      & & MPQ-DM~\cite{feng2025mpq}          & 4/4   & 76.35  & 22.35 & 35.18 & 0.5978 \\
      & & Baseline~\cite{yang2025robuq}   & 4/4   & 192.13 & 9.87  & 25.05 & 0.7322 \\
      & &  \textbf{TreeQ-PEFT} &  4/4 &  \textbf{206.10} &  \textbf{8.69} &  \textbf{23.15} &  \textbf{0.7449} \\
      \cdashline{3-8}
      & & MPQ-DM~\cite{feng2025mpq}          & 3/3   & 3.45   & 234.69 & 89.27 & 0.1137 \\
      & & Baseline~\cite{yang2025robuq}   & 3/3   & 128.06 & 16.63 & 29.20 & 0.6745 \\
      & &  \textbf{TreeQ-PEFT} &  3/3 &  \textbf{135.84} &  \textbf{14.52} &  \textbf{26.49} &  \textbf{0.6869} \\
    \midrule

    \multirow{13}{=}{\centering ImageNet\\CFG = 4.0\\steps = 50}
      & \multirow{7}{=}{\centering PTQ} 
      & FP               & 32/32 & 478.35 & 19.11 & 21.61 & 0.9298 \\
      \cdashline{3-8}
      & & SVD-Quant~\cite{li2024svdquant}       & 4/4   & 6.86   & 164.82 & 68.17 & 0.0413 \\
      & & Quarot~\cite{ashkboos2024quarot}          & 4/4   & 375.95 & \textbf{13.05} & 24.14 & 0.8178 \\
      & & Baseline~\cite{yang2025robuq}           & 4/4   & 456.58 & 16.13 & 20.34 & 0.9291 \\
      & &  \textbf{TreeQ-PTQ} &  4/4 &  \textbf{460.35} &  17.33 &  \textbf{20.13} &  \textbf{0.9330} \\
      \cdashline{3-8}
      & & SVD-Quant~\cite{li2024svdquant}      & 3/3   & 1.60   & 300.07 & 215.19 & 0.0063 \\
      & & Quarot~\cite{ashkboos2024quarot}          & 3/3   & 4.51   & 273.61 & 173.75 & 0.0490 \\
      & & Baseline~\cite{yang2025robuq}           & 3/3   & 254.97 & 14.00 & 30.93 & 0.7160 \\
      & &  \textbf{TreeQ-PTQ} &  3/3 &  \textbf{329.62} &  \textbf{9.91} &  \textbf{20.84} &  \textbf{0.7943} \\
      \cline{2-8}
      & \multirow{4}{=}{\centering PEFT}
      & Baseline~\cite{yang2025robuq}   & 4/4   & 459.80 & 17.91 & \textbf{19.97} & 0.9297 \\
      & &  \textbf{TreeQ-PEFT} &  4/4 &  \textbf{470.10} &  \textbf{17.89} &  20.08 &  \textbf{0.9303} \\
      \cdashline{3-8}
      & & Baseline~\cite{yang2025robuq}   & 3/3   & 404.12 & \textbf{14.49} & 19.83 & 0.9180 \\
      & &  \textbf{TreeQ-PEFT} &  3/3 &  \textbf{413.45} &  15.03 &  \textbf{19.70} &  \textbf{0.9234} \\
    \bottomrule[0.15em]
  \end{tabular}
  }
  \vspace{-5mm}
\end{table*}
\vspace{-3mm}
\section{Experiments}
\vspace{-1mm}
\subsection{Setup}
\label{sec:setup}
\vspace{-1mm}

\noindent{\textbf{Datasets and Evaluation.}} We evaluate pre-trained class-conditional DiT-XL/2 models at $256{\times}256$ resolution on ImageNet-1K~\citep{russakovsky2015imagenet}. The DDPM solver~\citep{ho2020denoising} with 50 sampling steps is employed for generation. To comprehensively assess the generated images, we use four widely adopted metrics: Fr\'echet Inception Distance (FID)~\citep{heusel2017gans}, spatial FID (sFID)~\citep{salimans2016improved, nash2021generating}, Inception Score (IS)~\citep{salimans2016improved, barratt2018anote}, and Precision, all computed with the ADM toolkit~\citep{dhariwal2021diifusion}.

\noindent{\textbf{Compared Methods.}} We compare our TreeQ series with \textbf{several leading} state-of-the-art quantization approaches, covering both Post-Training Quantization (PTQ) and Parameter-Efficient Fine-Tuning (PEFT) paradigms. For \textbf{PTQ} comparison, we include general-purpose methods SVD-Quant~\citep{li2024svdquant} and Quarot~\citep{ashkboos2024quarot}, and DiT-specific methods Q-DiT~\citep{chen2024qdit} and PTQ4DiT~\citep{wu2024ptq4dit}. For \textbf{PEFT} comparison, we include QueST~\citep{wang2025quest} for low-bit PEFT quantization and MPQ-DM~\citep{feng2025mpq} for mixed-precision quantization. A strong RobuQ~\citep{yang2025robuq} baseline is included for reference as well.

\noindent{\textbf{Training and Quantization Details.}} All experiments are conducted with PyTorch~\citep{paszke2019pytorch} on a single NVIDIA RTX A6000-48GB GPU. For PEFT-based (QAT) methods, we employ the QLoRA~\citep{dettmers2023qlora} strategy with the AdamW optimizer~\citep{loshchilov2017decoupled} at a learning rate of $10^{-5}$ and zero weight decay. We train for 20k steps with a batch size of 24, setting QLoRA rank to 16. For methods with parallel branches (i.e., baseline RobuQ and our TreeQ), the full-precision branch is trained concurrently, as added VRAM cost and training latency are negligible. \textbf{For fair comparison, we employ per-channel weight/activation quantizers for all methods.} We keep embedding and final layers in full precision across all methods and maintain 8-bit precision for activation-activation matrix multiplication operations, which constitute minimal computation but exhibit high quantization sensitivity and instability.

\noindent{\textbf{Mixed-Precision Search Configuration.}}
Our search space targets the 28 blocks within DiT-XL/2. We focus on layers accounting for most FLOPs: \texttt{qkv}, \texttt{proj}, \texttt{fc1}, and \texttt{fc2}. The TSS merge queue initializes following this order. We exclude \texttt{adaln} layers from search due to negligible computational cost, assigning them uniform bit-width matching the target average. Candidate bit-widths are \{2, 3, 4, 5\}. For performance indication, we sample 64 training examples at random timesteps, perform forward passes with quantized models, and compute the MSE between their outputs and the corresponding full-precision outputs.

\subsection{Main Results}
\vspace{-2mm}
\textbf{Quantitative Results.} We present class-conditional generation results on ImageNet 256x256 in Tab.~\ref{tab:main_experiments_final}, comparing TreeQ with its baseline under PTQ and PEFT.

\noindent{\textbf{Post-Training Quantization.}}
TreeQ establishes new state-of-the-art PTQ results across bit-widths and guidance scales. At CFG = 1.5, TreeQ achieves an FID of \textbf{6.91} in W4A4 configuration, approaching the full-precision baseline (6.28) while \textbf{surpassing all existing 4-bit PEFT methods} (9.87). Its robustness is particularly evident in extreme W3A3 quantization, maintaining 23.21 FID where conventional PTQ fails catastrophically (99.99). Under CFG = 4.0, TreeQ demonstrates superior and consistent stability, attaining \textbf{9.90 FID} at W3A3 with markedly better spatial consistency (sFID 20.83 versus 30.93).

\noindent{\textbf{Parameter-Efficient Fine-Tuning.}}
When integrated with QLoRA, TreeQ delivers consistent improvements across all settings. At CFG = 1.5, it reduces W4A4 FID from 9.87 to \textbf{8.68} and W3A3 FID from 17.08 to \textbf{16.63}. Under CFG = 4.0, TreeQ maintains strong competitiveness with W4A4 FID of 17.32 (versus 17.91 baseline) while preserving excellent stability. These results validate that TreeQ, as a versatile and highly effective quantization framework, enhances performance across both PTQ and PEFT scenarios.

\subsection{Ablation Study}
\vspace{-1mm}
This section presents ablation studies on TSS and GMB to evaluate key design components. Experiments were performed on ImageNet-256 to test PTQ performance. We additionally designed text-to-image (T2I) and super-resolution (SR) tasks for GMB, successfully validating its effectiveness in preserving high-frequency information.

\subsubsection{Ablation on TSS}
\noindent{\textbf{Max Length.}}
We conducted an ablation study on the maximum length $k$ of TSS's Pareto queue to evaluate its PTQ performance impact. As shown in Tab.~\ref{tab:ablation_max_length}, we tested $L$ values of $\{4, 8, 12, 16, 32\}$. Given quadratic time complexity relative to $L$, \textbf{$L=16$} provides an optimal trade-off, achieving performance comparable to $L=32$ with only a quarter of the computational cost. Notably, since the candidate bit-width set size is 4, whose square equals $L$, first-round Pareto merging requires no computation, thereby yielding additional and noticeable efficiency gains.

\vspace{-2mm}
\begin{table}[t]
  \centering
  \caption{Ablation Study on the maximum length of Pareto queue.}
  \label{tab:ablation_max_length}
  \vspace{-3mm}
  \setlength{\tabcolsep}{3.6mm}
  \renewcommand{\arraystretch}{1.1}
  \resizebox{\linewidth}{!}{
  \begin{tabular}{C{2cm} C{0.9cm} C{1.1cm} C{1.1cm} C{1.1cm} C{1.1cm} C{1.1cm}}
    \toprule[0.15em]
    Setting & L & IS$\uparrow$ & FID$\downarrow$ & sFID$\downarrow$ & P$\uparrow$ & Time$\downarrow$ \\
    \midrule
    \multirow{5}{=}{\centering CFG = 1.5 \\ W4A4\\ 10K Samples}
      & 4  & 192.35 & 8.31  & 22.26 & 0.7457 & 0.65h\\
      & 8  & 200.28 & 7.74  & 21.91 & 0.7476 & 2.58h\\
      & 12 & 198.34 & 8.16  & 22.11 & 0.7460 & 5.81h\\
      & \cellcolor{gray!50}16 & \cellcolor{gray!50}198.73 & \cellcolor{gray!50}7.99 
         & \cellcolor{gray!50}21.62 & \cellcolor{gray!50}0.7517 & \cellcolor{gray!50}5.16h\\
      & 32 & 199.24 & 7.99  & 21.98 & 0.7497 & 20.64h \\
    \midrule
    \multirow{5}{=}{\centering CFG = 1.5 \\ W3A3\\ 10K Samples}
      & 4  & 78.43  & 32.85 & 35.99 & 0.5086 & 0.65h\\
      & 8  & 80.28  & 29.27 & 30.37 & 0.5365 & 2.58h\\
      & 12 & 87.55  & 26.64 & 29.84 & 0.5466 & 5.81h\\
      & \cellcolor{gray!50}16 & \cellcolor{gray!50}94.55 & \cellcolor{gray!50}23.21 
         & \cellcolor{gray!50}27.79 & \cellcolor{gray!50}0.5670 & \cellcolor{gray!50}5.16h\\
      & 32 & 93.93 & 23.82 & 27.92 & 0.5720 & 20.64h\\
    \bottomrule[0.15em]
  \end{tabular}
  }
  \vspace{-5mm}
\end{table}

\noindent{\textbf{Comparison with Other Bit-width Allocation Schemes.}}
We compared different bit-width allocation schemes, testing uniform PTQ quantization and integer programming on calibration sets using L1, L2, and Hessian metrics alongside our TSS algorithm. Results in Tab.~\ref{tab:ablation_mp_methods} demonstrate that TSS significantly outperforms other integer programming methods based on local metrics. This highlights that fully considering inter-layer interactions helps avoid optimization bias and achieve substantial improvement.

\begin{table}[t]
  \centering
  \caption{Ablation on Mixed-Precision Search Algorithms. We use calibration set size $S{=}64$ for TreeQ and 256 for others.}
  \label{tab:ablation_mp_methods}
  \vspace{-3mm}
  \setlength{\tabcolsep}{3.8mm}
  \renewcommand{\arraystretch}{1.1}
  \resizebox{\linewidth}{!}{
  \begin{tabular}{C{2.0cm} C{2.3cm} C{1.1cm} C{1.1cm} C{1.2cm} C{1.1cm}}
    \toprule[0.15em]
    Setting & Method & IS$\uparrow$ & FID$\downarrow$ & sFID$\downarrow$ & P$\uparrow$ \\
    \midrule
    \multirow{6}{=}{\centering CFG = 1.5\\W3A3\\5K Samples}
      & baseline   & 20.59 &  103.67  & 83.15 & 0.2328\\
      & IP+L2      & 60.17  &  49.47  & 61.53 &  0.4466 \\
      & IP+Hessian & 59.27  &  50.28  &62.34 & 0.4396 \\
      & IP+L1      &55.42  & 53.15  & 63.94 &  0.4380 \\
      & \cellcolor{gray!50}TreeQ 
        & \cellcolor{gray!50}85.21 & \cellcolor{gray!50} 28.08 
        & \cellcolor{gray!50}28.97 & \cellcolor{gray!50} 0.5346 \\
      & FP  &241.17&	6.28&	20.78&	0.7830\\
    \bottomrule[0.15em]
  \end{tabular}
  }
  \vspace{-3mm}
\end{table}

\begin{table}[t]
  \centering
  \caption{Ablation study on calibration set size.}
  \label{tab:ablation_calibration_size}
  \vspace{-3mm}
  \setlength{\tabcolsep}{3.5mm}
  \renewcommand{\arraystretch}{1.1}
  \resizebox{\linewidth}{!}{
  \begin{tabular}{C{2.0cm} C{0.9cm} C{1.1cm} C{1.1cm} C{1.1cm} C{1.1cm} C{1.1cm}}
    \toprule[0.15em]
    Setting & S & IS$\uparrow$ & FID$\downarrow$ & sFID$\downarrow$ & P$\uparrow$ & iter/s$\uparrow$\\
    \midrule
    \multirow{6}{=}{\centering CFG = 1.5 \\ W4A4 \\ 5K Samples}
      & 4   & 168.35 & 15.26 & 42.12 & 0.7004 & 5.96\\
      & 8   & 178.95 & 14.38 & 40.98 & 0.7280 & 5.89\\
      & 16  & 180.87 & 13.11 & 40.26 & 0.7372 & 3.04\\
      & 32  & 184.24 & 12.94 & 39.84 & 0.7423 & 1.52\\
      & \cellcolor{gray!50}64  
        & \cellcolor{gray!50}194.96 
        & \cellcolor{gray!50}12.10 
        & \cellcolor{gray!50}39.23 
        & \cellcolor{gray!50}0.7494 
        & \cellcolor{gray!50}0.76 \\
      & 256 & 192.24 & 12.99 & 40.13 & 0.7374 & 0.19\\
    \bottomrule[0.15em]
  \end{tabular}
  }
  \vspace{-5mm}
\end{table}
 
\noindent{\textbf{Calibration Set Size.}}
We further conducted an ablation study on calibration set size to evaluate quantization performance impact. As shown in Tab.~\ref{tab:ablation_calibration_size}, progressively increasing calibration set size yields significant performance gains. We adopt a \textbf{calibration set size of 64} for optimal performance while maintaining computational efficiency.

\noindent{\textbf{Analysis of Saturation.}}
We observed that performance does not consistently improve with larger candidate sets or queue lengths, primarily because the MSE optimization objective is not perfectly order-preserving with final metrics. However, balancing search speed and performance, we find MSE a strongly order-preserving proxy sufficiently effective. Our method can be enhanced by employing metrics with better order-preserving properties.
\begin{table}[th]
 \vspace{-3mm}
  \centering
  \caption{Ablation studies on GMB design. 10K samples evaluated with CFG=4.0 at average W4A4.}
  \label{tab:ablation_GMB_design}
  \vspace{-3mm}
  \setlength{\tabcolsep}{3.2mm}
  \renewcommand{\arraystretch}{1.1}
  \resizebox{\linewidth}{!}{
  \begin{tabular}{C{2.0cm} C{2.3cm} C{1.1cm} C{1.1cm} C{1.1cm} C{1.1cm}}
    \toprule[0.15em]
    Ablation & Method & IS$\uparrow$ & FID$\downarrow$ & sFID$\downarrow$ & P$\uparrow$ \\
    \midrule
    \multirow{2}{=}{\centering Init. Order}
      & GMB-first & 195.71 & 8.32 & 22.57 & 0.7478 \\
      & \cellcolor{gray!50}LRB-first 
        & \cellcolor{gray!50}208.77 
        & \cellcolor{gray!50}7.53 
        & \cellcolor{gray!50}21.68 
        & \cellcolor{gray!50}0.7527 \\
    \midrule
    \multirow{2}{=}{\centering GMB Pos.}
      & Pre-Hadamard & 208.77 & 7.53 & 21.68 & 0.7527 \\
      & \cellcolor{gray!50}Post-Hadamard 
        & \cellcolor{gray!50}219.66 
        & \cellcolor{gray!50}6.92 
        & \cellcolor{gray!50}20.86 
        & \cellcolor{gray!50}0.7664 \\
    \midrule
    \multirow{4}{=}{\centering Part Num.}
      & $r = 0$  & 208.36 & 7.46 & 21.54 & 0.7553 \\
      & \cellcolor{gray!50}$r = 4$  
        & \cellcolor{gray!50}219.66 
        & \cellcolor{gray!50}6.92 
        & \cellcolor{gray!50}20.86 
        & \cellcolor{gray!50}0.7664 \\
      & $r = 8$  & 214.29 & 6.90 & 20.47 & 0.7643 \\
      & $r = 16$ & 212.35 & 7.13 & 20.76 & 0.7735 \\
    \bottomrule[0.15em]
  \end{tabular}
  }
  \vspace{-4mm}
\end{table}

\vspace{-1mm}
\begin{figure*}[t!]
 \vspace{-4mm}
 \centering
 \includegraphics[height=6cm]{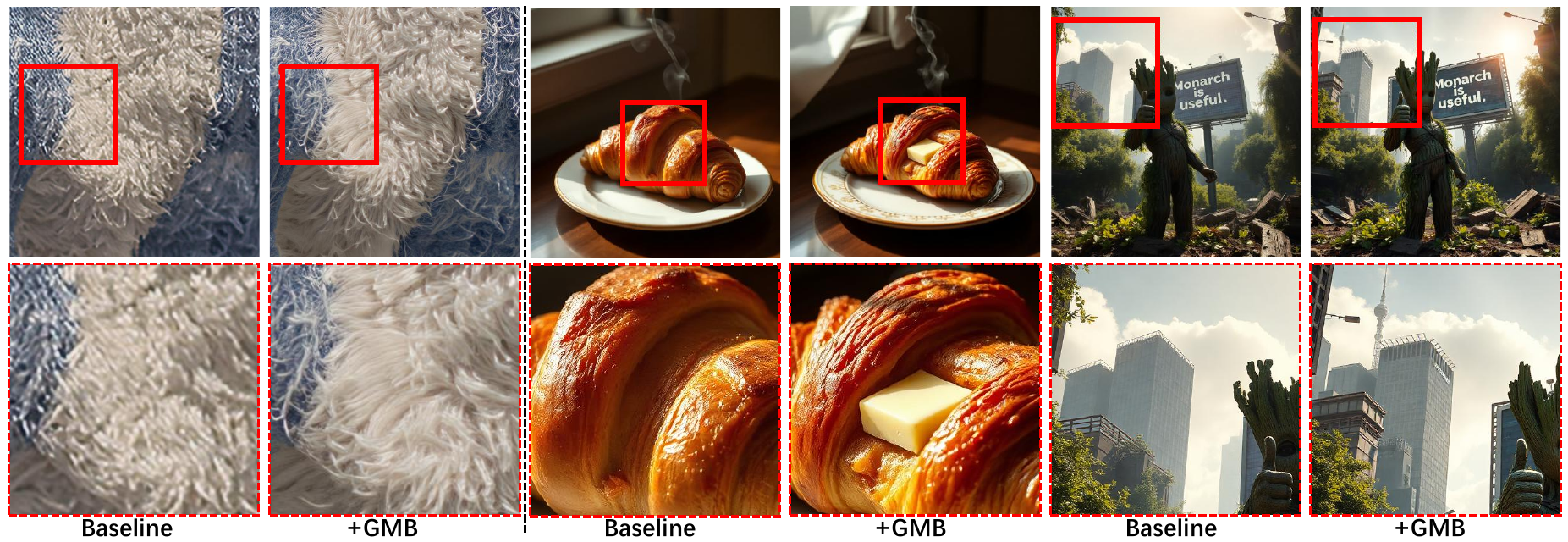}
 \vspace{-4mm}
 \caption{GMB provides more details for low-bit quantized DiT4SR (left) and FLUX-Schnell (right), advancing practical applications.}
 \vspace{-7mm}
 \label{fig:visual_compare_monarch}
\end{figure*}

\subsubsection{Ablation on GMB}
\vspace{-1mm}
\noindent{\textbf{Initialization Order.}}
We ablate two initialization strategies: (i) initializing GMB before the low-rank branch (LRB), and (ii) initializing LRB first from the Hadamard-transformed weight~$\mathbf{W}_H$, then deriving GMB from the residual. Results in Tab.~\ref{tab:ablation_GMB_design} show that \textbf{(ii)} yields superior performance, indicating LRB more effectively absorbs the dominant low-rank principal components, allowing GMB to better model the remaining high-rank residuals.

\noindent{\textbf{Placement w.r.t. Hadamard.}}
We study whether GMB should be placed before or after the Hadamard transform. Experimental results favor post-transform placement. We attribute this to the Hadamard transform's whitening effect, which promotes a more uniform weight distribution and further enables subsequent GMB to effectively retain more informative components during quantization.

\noindent{\textbf{Partition Number.}}
We experiment with GMB's partition number $r$. Using LRB with rank 16, we observe that setting \textbf{partition number $r=4$} yields significant performance improvements and enhanced visual details while introducing minimal additional parameters or computational overhead.
\noindent{\textbf{Ablation on High-Detail Tasks.}}
To further validate GMB's generalizability and model-agnostic capabilities, we conducted supplementary experiments on high-detail tasks. For SR task, we tested on the DiT4SR~\cite{duan2025dit4sr} framework using the RealSR~\cite{cai2019realsr} dataset. As shown in Tab.~\ref{tab:realsr_w4a6}, our GMB inclusion consistently improves a suite of key perceptual metrics, including MUSIQ~\cite{ke2021musiq}, MANIQA~\cite{yang2022maniqa}, and LPIPS~\cite{zhang2018lpips}. For T2I task, we evaluated on the FLUX-Schnell~\cite{blackforestlabs2024flux} model using the MJHQ-5k~\cite{zhang2023mjhq} dataset. Tab.~\ref{tab:flux_w4a4} shows a similar trend, where GMB improves the FID~\cite{heusel2017gans} and significantly boosts perceptual scores like ImageReward~\cite{xu2023imagereward}. The qualitative results in Fig.~\ref{fig:visual_compare_monarch} also corroborate that GMB restores finer details. This confirms GMB's effectiveness in preserving high-frequency fidelity across different tasks. More detailed experiment settings are given in Supplementary Material.

\begin{table}
  \centering
  \caption{Results on RealSR under W4A6 setting.}
  \vspace{-4mm}
  \label{tab:realsr_w4a6}
  \setlength{\tabcolsep}{3.6mm}
  \renewcommand{\arraystretch}{1.15}
  \resizebox{\linewidth}{!}{
  \begin{tabular}{lccccc}
    \toprule[0.15em]
    \textbf{Method} & LPIPS$\downarrow$ & MUSIQ$\uparrow$ & MANIQA$\uparrow$ & ClipIQA$\uparrow$ & NIQE$\uparrow$ \\
    \midrule
    FP & 0.3190 & 68.07 & 0.6610 & 0.5503 & 3.977 \\
    \cdashline{1-6}
    Baseline & \textbf{0.3248} & 65.29 & 0.4249 & 0.6212 & 3.516 \\
    \textbf{Baseline+GMB} &0.3364 &\textbf{67.10} & \textbf{0.4339} & \textbf{0.6439} &\textbf{3.788} \\
    \bottomrule[0.15em]
  \end{tabular}}
   \vspace{-4mm}
\end{table}
\begin{table}
  \centering
  \caption{Results on MJHQ-5k under W4A4 setting.}
   \vspace{-4mm}
  \label{tab:flux_w4a4}
  \setlength{\tabcolsep}{3.6mm}
  \renewcommand{\arraystretch}{1.15}
  \resizebox{\linewidth}{!}{
  \begin{tabular}{lccccc}
    \toprule[0.15em]
    \textbf{Method} & FID$\downarrow$ & Image Reward$\uparrow$ & CLIPIQA$\uparrow$ & CLIPScore$\uparrow$ & PSNR$\uparrow$ \\
    \midrule
    FP & 18.40 & 0.9323 & 0.9399 &  26.54 & - \\
    \cdashline{1-6}
    Baseline & 18.57 & 0.8617 & 0.9266 & \textbf{26.34}& 17.21 \\
    \textbf{Baseline+GMB} & \textbf{18.37} &\textbf{0.9035} & \textbf{0.9402} &  26.29  & \textbf{17.27} \\
    \bottomrule[0.15em]
  \end{tabular}}
  \vspace{-5mm}
\end{table}

\vspace{-3mm}
\section{Limitations and Future Work}
\vspace{-2mm}
\noindent\textbf{Lack of Joint Optimization.}
A limitation of our current work is the absence of a unified framework for jointly optimizing quantization, low-rank decomposition, and structured sparsity. Our proposed GMB branch (Sec.~\ref{sec:GMB}) serves as an effective method for introducing sparsity. However, while it has demonstrated considerable potential in its current application, we have not yet fully explored the theoretical aspects of optimizing all three components simultaneously. This remains a key direction for future research.
\noindent\textbf{Support for Diverse Quantization Kernels.}
We have successfully implemented a dedicated kernel for our method in the W4A4 configuration, achieving significant gains. However, similar to challenges faced by the broader research community, efficient and publicly available implementations for other ultra-low-bit kernels—such as W2A2 and W3A3—remain scarce. We plan to develop these additional kernels in future to achieve better acceleration and demonstrate the full hardware potential of our approach.
\vspace{-1mm}
\section{Conclusion}
\vspace{-2mm}
In this paper, we propose \textbf{TreeQ}, a novel and comprehensive framework addressing the challenging performance degradation in ultra-low-bit Diffusion Transformers (DiTs). Our framework introduces three critical contributions: \textbf{Tree-Structured Search (TSS)}, which leverages DiT's local dependencies for efficient topology-aware bit configuration; \textbf{Environmental Noise Guidance (ENG)}, which unifies the distinct PTQ and QAT objectives via a simple hyperparameter; and the \textbf{General Monarch Branch (GMB)}, a parallel branch that recovers high-frequency details during quantization. Extensive experiments demonstrate new state-of-the-art performance. Notably, our 4-bit PTQ achieves an FID of 6.92, remarkably close to the full-precision model and even surpassing strong PEFT baselines. This promising advancement, achieved while preserving high-frequency detail integrity, represents a significant step toward the practical deployment of low-bit DiTs.

\clearpage
\clearpage
{
    \small
    \bibliographystyle{ieeenat_fullname}
    \bibliography{main}
}

\end{document}